\documentclass{article}
\pdfoutput=1
\usepackage{spconf,amsmath,graphicx}
\usepackage[utf8x]{inputenc}
\usepackage[OT1]{fontenc}

\usepackage{amsfonts}
\usepackage{amssymb}
\usepackage{stmaryrd}
\usepackage{hyperref}

\usepackage{pgf}

\usepackage{xargs}                      % Use more than one optional parameter in a new commands
\usepackage[colorinlistoftodos,prependcaption]{todonotes}
\newcommandx{\unsure}[2][1=]{\todo[linecolor=red,backgroundcolor=red!25,bordercolor=red,#1]{#2}}
\newcommandx{\change}[2][1=]{\todo[linecolor=blue,backgroundcolor=blue!25,bordercolor=blue,#1]{#2}}
\newcommandx{\info}[2][1=]{\todo[linecolor=OliveGreen,backgroundcolor=OliveGreen!25,bordercolor=OliveGreen,#1]{#2}}
\newcommandx{\improvement}[2][1=]{\todo[linecolor=Plum,backgroundcolor=Plum!25,bordercolor=Plum,#1]{#2}}
\def\X{{\mathbf X}}
\def\V{{\mathbf V}}
\def\U{{\mathbf U}}
\def\E{{\mathbf E}}
\def\P{{\mathbf P}}
\def\R{{\mathbb{R}}}
\def\Y{{\mathbf Y}}
\def\Q{{\mathbf Q}}
\def\v{{\mathbf v}}

\def\bigO{\mathcal{O}}

\newcommand{\transpose}{^\mathsf{T}}

\newcommand{\argmin}{\operatornamewithlimits{argmin}}

\title{Compressed Online Dictionary Learning for Fast Resting-State fMRI Decomposition}

\name{Arthur Mensch$^{(1)}$ \qquad Gaël Varoquaux$^{(1)}$ \qquad Bertrand Thirion$^{(1)}$}

\address{$^{(1)}$Parietal team, Inria, CEA, Paris-Saclay University. Neurospin, 91191 Gif-sur-Yvette, France}

\begin{document}
\ninept
\maketitle
\begin{abstract}We present a method for fast resting-state fMRI spatial
decompositions of very large datasets, based on the reduction of the
temporal dimension before applying dictionary learning on concatenated %  \textit{reduced}
% bertrand: reduction is already in the sentence. I'm afraid to create some confusion.
individual records from groups of subjects. Introducing a measure of
correspondence between spatial decompositions of rest fMRI, we
demonstrates that time-reduced dictionary learning produces result as
reliable as non-reduced decompositions. We also show that this reduction
significantly improves computational scalability.
\end{abstract}
\section{Introduction}
\label{sec:intro}

Resting-state fMRI data analysis traditionally implies, as an initial step, to decompose a set of raw 4D records (time-series sampled in a volumic voxel grid) into a sum of spatially located \textit{functional networks} that isolate a part of the brain signals.
Functional networks, that can be seen as a set of brain \textit{activation maps}, form a relevant basis for the experiment signals that captures its essence in a low-dimensional space. As such, they have been successfully used for feature extraction before statistical learning, \textit{e.g.} in decoding tasks.

While principal component analysis (PCA) on image arrays has been the first method to be proposed for fMRI, independent component analysis (ICA) is presently the most popular decomposition technique in the field. It involves finding a spatial basis $\V$ that is closest to a set of spatially \textit{independent} sources.
More recent work have shown that good results can be obtained imposing
\textit{sparsity} rather than \textit{independence} to spatial
decomposition \cite{varoquaux_multisubject_2011}, relying on
\textit{dictionary learning} formulation
\cite{kreutz-delgado_dictionary_2003}.

All these techniques suffer from their lack of scalability, as they were initially designed to be applied to small datasets.
The recent increase in publicly available dataset size (\textit{e.g.} HCP \cite{vanessen_human_2012}) has revealed their limits in terms of memory usage and computational time.
Efforts have been made to make decomposition methods available for large scale studies, possibly with several groups.
They involve using a hierarchical model for dictionary learning \cite{varoquaux_multisubject_2011} or incremental PCA techniques \cite{smith_grouppca_2014}.
However, the former only proposes PCA+ICA based decomposition methods, which do not naturally yield sparse maps, and the latter suffers from its computational complexity.
Running a satisfying decomposition algorithm on the full HCP dataset
currently requires a very large workstation.

In this paper, we focus on dictionary learning methods for fMRI, and show
how to make them more scalable in both time and memory. Uncovering the
computational limitations of dictionary learning when analysing very
large datasets, we propose to perform random-projection based
hierarchical dimension reduction in the time direction before applying
dictionary learning methods. As a result, time and memory consumption are reduced, avoiding out-of-core computation.
We introduce a measure of correspondence to relate results obtained
from compressed data to those from non compressed data, and show that
substantial gain in time and memory can be obtained with no significant
loss in quality of the extract networks.
%Secondly, we transpose the neuro-imaging formulation of dictionary learning so that it becomes tractable and out-of-core performable, using online sparse PCA methods decribed in \cite{mairal_online_2010}. We use our validation framework to show the accuracy of our results compared to existing techniques.

\section{Scalabity of dictionary learning for fMRI}

\subsection{rfMRI decomposition existing formalism}
We consider multi-subject rfMRI data: a set of matrices $(\X^s)_{s \in [1, t]}$ in $(\R^{n \times p})^t$, with $p$ voxels per volume, $n$ temporal samples per record, and $t$ records.
We seek to decompose it as :
\begin{align}
\forall s \in \left\llbracket1, t\right\rrbracket\!,\,\X^s = \U^s
\V\transpose\quad\text{with}\;\;\U^s \in \R^{n \times k}\!,\; \V \in \R^{p \times k}
\end{align}
Existing decomposition techniques vary in the criterion they optimize, and on the hierarchical model they propose.
We focus on dictionary learning methods, that have been shown to obtain better results than ICA in \cite{varoquaux_multisubject_2011}.
To handle group studies, we choose the most simple hierarchical model, that consists in performing time concatenation of the records -- first proposed by \cite{calhoun2001method} for ICA.
We write $\U \in \R^{nt \times k}$ and $\X \in \R^{nt \times p}$ the vertical concatenation of $(\U^s)_s$ and $(\X^s)_s$, and seek to decompose $\X$ instead of $\X^s$.

A good decomposition should allow a good reconstruction of data while
being spatially localized, i.e. \textit{sparse} in voxel space. Such a
decomposition setting can be formalized in a \textit{dictionary learning}
(DL) optimization framework, that combines a sparsity inducing penalty to
a reconstruction loss. We seek to find $k$ dense \textit{temporal} atoms, \textit{i.e.}
time-series, that will constitute loadings for $k$ sparse spatial maps
with good signal recovery. In one of its original formulation
\cite{kreutz-delgado_dictionary_2003}, this leads to the following
optimization problem:
\begin{equation}
\label{eq:dl}
\min_{\substack{\U \in \R^{nt \times k},\\
\V \in \R^{p \times k}}}
\Vert \X - \U \,\V\transpose \Vert_F^2 + \lambda \left\Vert \V \right\Vert_1\,\text{s.t.}\,\forall j, \left\Vert \U_j \right\Vert_2 \leq 1
\end{equation}
Each row $L_i(\V)$ yields the sparse $k$ loadings related to the $k$ temporal atoms for a single voxel time-serie, held in column $X^i$.
\cite{mairal_online_2010} introduces an efficient online solver
for this minimization problem, streaming on \textit{voxel time-series},
i.e loading $\X$ columnwise: at iteration $t$, a voxel time-serie batch
$L_{b(t)}(\V)$ is computed (using a Lasso solver) on the present dictionary $\U_{t-1}$, and $\U_{t}$ is updated (using block coordinate descent) to best reconstruct previously seen time-series from previously computed sparse codes.
The final spatial components are then obtained solving Lasso problems
$\min_{ \V \in \R^{p \times k}}\Vert \X - \U_{\textrm{end}} \V\transpose \Vert_F^2 + \lambda \Vert \V \Vert_1$.

This online algorithm provably converges towards a solution of Eq.\;\ref{eq:dl}
under conditions satisfied in neuro-imaging. A good initialization for
temporal atoms is required to obtain an exploitable solution. It can
typically be obtained by computing time-series associated to an initial
guess on activation maps $\V_{\mathrm{init}}$, \textit{e.g.} obtained from known brain networks. The
temporal atoms are computed by
solving $\min_{\U_i} \Vert \X_i - \U_i
\V_{\textrm{init}}\transpose \Vert_2$ for all $i \in \llbracket 1,n \rrbracket$.

%solving:
%\begin{equation}
%\min_{\U \in \R^{nt \times k}} \frac{1}{t} \sum_{i=1}^t \Vert X^i - \U(t) \V_i^T \Vert_F^2 + \lambda \Vert \V_i^T \Vert_1
%\end{equation}

\subsection{Scalability challenge}

%\paragraph*{Theoretical scalability}Following \cite{mairal_online_2010}, adding time complexity of Lasso solver and block coordinate descent, each iteration has a complexity of $O (bnk^2+bks^2)$, with $b$ batch size and $s < k$ number of active components in each $\V$ rows. Convergence is typically reached in one epoch, leading to a theoretical complexity of $O(pnk^2)$, and spatial complexity of $O(n(b+k))$.
Following \cite{mairal_online_2010}, online dictionary learning has an
overall complexity of $\bigO(n\,p\,k^2)$, as convergence is typically reached
within one epoch on rfMRI. In theory, the dictionary learning problem is
thus computationally scalable. However, on large rfMRI datasets,
online dictionary learning faces two main challenges detailed below.

\paragraph*{Out-of-core requirements for large datasets}For datasets like
HCP ($t\!=\!2000$, $n\!=\!1200$, $p\!=\!20000$, $1.92\mathrm{TB}$),
typical computers are unable to hold all data in memory.
It is thus necessary to stream the data from disk, which is only reasonably efficient if the data are stored in the same direction as it is accessed.
Yet online DL algorithm require to pass 3 times over data, during which it is streamed in different directions (row-wise for initialization, columnwise for DL and final Lasso solving), while fMRI images are naturally stored row-wise.
For the sake of efficiency, storage copy and manipulation is required, which is a serious issue for neuroscientists dealing with over $1\textrm{TB}$ datasets. Going out-of-core sets a large \textit{performance gap} between small datasets and large datasets.

\paragraph*{Grid search in parameter setting} The sparsity of the maps obtained depends critically on parameter $\lambda$, that scales non trivially with $p$. It is therefore impossible to set it independently from the \textit{experiment size}, and several runs must be performed to obtain best maps, relative to their neurological relevance or a validation criterion.
Grid search should be run in parallel for efficiency, which is a serious issue when doing out-of-core computation, as simultaneous access to the disk from different processes makes the pipeline IO-bound.
Reducing dataset size therefore reduces disk and memory usage, which permits the efficient use of more CPUs.

\medskip
Both issues suggest to reduce memory usage by reducing datasets size while keeping the essential part of its signal: being able to keep data in memory avoids drastic loss in performance.

\section{Time-compressed dictionary learning}

\paragraph*{Reducing time dimension}Good quality maps are already obtained using small datasets with standard number of samples (ADHD dataset, $n\!=\!150$).
For this reason, we investigated how large datasets can be reduced to fit in memory while keeping reasonable map accuracy compared to the non-reduced version.

Indeed, the $n$ time samples per subject are not uniformly scattered in voxel space, and should exhibit some low dimension structure: we expect them to be scattered close to some low rank subspace of $\R^p$, spanned by a set of $m$ vector $\X_r^s \in \R^{m \times p}$. We thus perform a \textit{hierarchical} rank reduction : $\X^s$ is first approximated by a rank $m$ surrogate $\P\transpose \X_r^s$, and a final rank $k$ decomposition is computed over concatenated data. We show that such reduction is conservative enough to allow good map extraction. Geometrically, we project $\X$ on a low rank subset of $\R^{n \times p}$:
\begin{align}
\label{eq:rank_red}
\P = \argmin_{\Q \in \R^{n\times m}} \left\Vert \X^s - \Q\,\Q\transpose\X^s \right\Vert_F \quad
\X_r^s = \P\transpose \X^s
\end{align}
Then $\X^s = \P\,\X_r^s + \E^s$ where $\E^s$ is a residual full rank noise matrix.

We approximate $\X^s$ with $\X_r^s$ at \textit{subject} level to retain
subject variability. Hence, replacing $\X$ with $\X_r$, the concatenation of
$(\X_r^s)$, in Eq.\;\ref{eq:dl}, we obtain a \textit{reduced} dictionary learning objective.
% stubbed \stepcounter{equation} (\label{eq:reduced}\theequation).

Importantly, we must have $m\,t{>}k$ so that $\X_r$ is at least of rank
$k$ to recover $k$ sparse activation maps. On the other hand, we show
that reducing $\X_r^s$ matrix beyond $m{<}k$ can still provide good results.

In our reduced dictionary learning algorithm, time and memory complexity are reduced by a factor $\alpha\!=\!\frac{m}{n}$, where $m$ should typically be of the same order than $k$.
%For example, setting $m = 120$ and $k = 70$, this allows, for typical large rfMRI datasets, to go from out-of-core size (2TB) to memory loadable size (200GB).%
This linear speed-up becomes much more dramatic when reduction allows to go from out-of-core to in-core computation.
It comes to the cost of the time required for matrix reduction that we
study in the following paragraph.

While Eq.\;\ref{eq:rank_red} can be seen as another way of decomposing $(\X^s)_s$, let us stress that this decomposition is performed in voxel space, in contrast with  dictionary learning itself, that identify a good basis in \textit{time} space.
The objective is to quickly find a good \textit{summary} of each $(\X^s)_s$ prior to applying dictionary learning, so as to reduce the dimensionality of the dictionary learning problem.

\paragraph*{The range-finding approach}$\X_r^s$ can be computed exactly with truncated SVD, following Eckart–Young–Mirsky theorem.
%: we write $\A \mathbf\Sigma \B^T$ the SVD decomposition of $\X^s$, with $\A \in \R^{n \times p}$ and $\B \in \R^{p \times p}$. We have $\X_r^s = \A_m \mathbf \Sigma_m \B_m^T$, with $\A_m, \B_m$ matrices formed with the $m$ first columns of $\A, \B$.%
However, exact SVD computation is typically $\bigO(p\,n^2)$, which is above dictionary learning complexity and makes prior data reduction useless when trying to reduce \textit{both} computation time and memory usage.
Fortunately, we show that we do not need exact $m$ rank best
approximation of $\X$ to obtain a satisfying $\V$. Following
\cite{halko_finding_2009} formalism, we seek $(\hat \P^s)_s \in (\R^{n \times m})^t$ such that
\begin{align}
\Vert \X^s - \hat\P^s \,\hat\P^s\!\transpose \X^s \Vert_F &\approx
\hspace*{-1em}\min_{\mathrm{rank}(\Y^s) \leq m}\Vert \E^s \Vert = \Vert \X^s - \Y^s \Vert_F
\end{align}
In \cite{halko_finding_2009}, Alg. 4.4, Halko proposes a fast, randomized
algorithm to compute such $\hat \P^s$, with measurable precision
$\Vert \hat \E^s - \E^s \Vert$. Setting $\hat\P = \mathrm{Diag}((\hat
\P^s))$, $\hat \X_r = \hat\P \X$, we use this random range-finding (rf)
algorithm to solve Eq.\;\ref{eq:dl}, where we replace $\X$ with $\hat \X_r$:
\begin{equation}
\label{eq:approx}
\min_{\substack{\U_r \in \R^{mt \times k}\\ \V \in \R^{p \times k}}}
\left\Vert \hat \P \,\X - \U_r \V\transpose \right\Vert_F^2 + \lambda \left\Vert \V \right\Vert_1\,\text{s.t.}\,\left\Vert (\U_r)_j \right\Vert_2 \leq 1
\end{equation}

The randomized range finding algorithm has a complexity of $\bigO(n\,p\,m)$, which is of same order as dictionary learning algorithm.
In practice, we show in Sec.\;\ref{sec:results} that its cost becomes negligible with respect to the reduction of dictionary learning cost, when the reduction ratio is high enough.

In a more straightforward way, we can set $\X_r^s = \X^s_{I}$, with $I$ subset (ss) of $\llbracket 1, n \rrbracket$ of size $m$. This category of reduction includes time subsampling of records. In this case, $\Vert \hat \E^s_{\mathrm{ss}} - \E^s \Vert$ cannot be controlled, and is expected to be larger than $\Vert \hat \E^s_{\mathrm{rf}} - \E^s \Vert$. Subsampling, for example, is expected to alias high frequency signal in records, preventing the recovery of activation maps with high frequency loadings in final dictionary learning application.

%In Sec. \ref{sec:validation}, we show how obtainable solution set $\lbrace\V_r\rbrace$ can be compared to solution set $\lbrace\V\rbrace$ obtainable from Eq. \ref{eq:dl} resolution.

\section{Validation}
\label{sec:validation}

\paragraph*{Reference result-set} Validation of dictionary learning
methods for rfMRI is challenging, as there is no ground truth to assess the quality of resulting map sets.
However, we can assess how much a result-set $\V$ obtained on a reduced
dataset $\X_r$ from Eq.\;\ref{eq:approx} is comparable to a result-set
$\V^0$ obtained on $\X$ from Eq.\;\ref{eq:dl}.

%We write $\V_i = \mathrm{DL}^0_i(\X)$ the results obtained from a run $i \in \mathbb N$ of DL over non reduced dataset $\X$.

%For reference, we choose $i < 0$ as we we will then compare $\V_i^0$ to $\V_i = \math

%Activation maps $\V_r$ solving approximate Eq. \ref{eq:reduced} have the same dimension as mpas $\V$ solving Eq. \ref{eq:dl}. We seek to design a similarity metric $d(\V, \V_r)$ where $d \sim 1$ when $\V$ and $\V_r$ are \textit{comparable}, in a sense to be defined, and $d \sim 0$ comparing two random sets.

\paragraph*{Result-set comparison}Two sets of maps $\V^0$ and $\V$ can
only be compared with an indicator invariant to map ordering. Two sets are
\textit{comparable} if each map from the first set is comparable to a map
in the second set. We find the best one-to-one coupling between these two
sets of maps and compute correlation between each \textit{best assigned}
couple of maps: $\mathrm{corr}(\v^0_i, \v_j) = \frac{\vert
(\v^0_i)\transpose \v_j \vert }{\Vert \v^0_i \Vert_2 \Vert \v_j \Vert_2}$ to measure similarity between two maps $\v_j$ (held in column $C_j(\V)$) and $\v_i^0$. We set $d$ to be the mean correlation between best assigned maps:
\begin{equation}
d(\V, \V^0) = \max_{\mathbf \Omega \in \mathcal{S}_k} \mathrm{Tr}
\bigl(\V\transpose \mathbf \Omega \,\V^0\bigr)
\end{equation}
where $S_k$ is the set of permutation matrices. $\mathbf \Omega$ can be computed efficiently using the Hungarian algorithm.

\paragraph*{Comparing random results}Eq.\;\ref{eq:dl} and \ref{eq:approx} admits many local minima that depend on algorithm initialization, and on the order used for streaming dataset columns. For any dataset $\Y \in \lbrace \X, (\X_r)_\mathrm{method}^\mathrm{reduction} \rbrace$, we expect obtained maps $\V_i = \mathrm{DL}_i (\Y)$ to capture a neurological/physical phenomenon for any run $i$ corresponding to a streaming order.
As in \cite{himberg_validating_2004}, we perform $l$ runs numbered on $\mathcal S_l \subset \mathbb{N}$ of the algorithm to obtain different maps, and compare the \textit{concatenation} $\mathcal{V}_l (\Y) = [ (\V_i)_{i \in \mathcal S_l} ]$ of these maps to the concatenation of reference maps $\mathcal{V}^0_l (\X) = [ (\V^0_i)_{i \in \mathcal S^0_l}]$ with runs numbered on $\mathcal S_l^0$:
\begin{equation}
d_l(\X, \mathcal S^0_l, \Y, \mathcal S_l) = d\bigl(\mathcal V_l^0(\X),
\mathcal V_p(\Y)\bigr)
\end{equation}
We thus take into account non unicity of DL solutions: different maps are obtained when performing the dictionary learning algorithm over the \textit{same} data with the \textit{same} parameters. We model result maps $(\v_i)_i$ to be part of a larger \textit{full} result-set $\mathcal{V}$:
\begin{align}
\label{eq:result_set}
\mathcal{V} (\Y)= \Big\lbrace
 \v_i = C_i (\V) \in \R^{p} \text{ s.t. }
 \V \in \R^{p \times k},
 \exists \U \in \R^{n \times k}&, \notag \\
 \left(\U, \V \right) \in \argmin_{\U, \V}
 \left\Vert \Y - \U \V\transpose
 \right\Vert_F^2
 + \lambda \left\Vert \V \right\Vert_1 &\Big\rbrace
\end{align}
When result-sets are concatenated over all possible streaming orders, we expect $d_p$ to converge toward a $\mathcal S_p^{(0)}$ independent measure:
\begin{align}
\label{eq:empirical}
&\phantom{=}d_\infty\bigl(\mathcal{V}(\X), \mathcal{V}(\Y)\bigr) = \lim_{l
\rightarrow \infty} d_l\bigl(\X, \mathcal S^0_l, \Y, \mathcal S_l\bigr)
\end{align}

It is expected that $d_\infty (\mathcal V (\X), \mathcal V (\X)) = 1$, but $p$ is finite in practice. Ensuring $\mathcal S^0_l \cap \mathcal S_l = \emptyset$, we measure mean result-set correspondence $d_l(\X, S_l^0, \X, S_l)$ over different runs on the same dataset $\X$, and compare it to $d_l(\X, S_l^0, \Y, S_l)$ to assess the \textit{reduction} effect.

%where $(\V_i)_i\!=\!\Vhatcal^p$, $((\V_r)_i)_i=\Vhatcal_r^p$ are obtained from various runs of respectively Eq. \ref{eq:dl} and \ref{eq:approx} resolutions and are concatenated columnwise. $\hat d_p$ is thus an empirical measure depending on $\X$, $\X_r$.

\section{Results}
\label{sec:results}

\paragraph*{Tools and datasets}We validate our reduction framework over two different datasets with different size: ADHD data, with 40 records, $n\!=\!150$ time steps per record; a subset of HCP dataset, using 40 subjects, 2 records per subject, subsampling records from $n\!=\!1200$ to $n\!=\!400$ to obtain reference $\X$.

Dictionary learning output depends on its initialization, and the problem of choosing the \textit{best} number of components $k$ is very ill-posed.
We bypass these problems by choosing $k\!=\!70$ for HCP dataset,
$k\!=\!20$ for ADHD dataset, and use reference ICA-based maps RSN20 and RSN70 from \cite{smith_correspondence_2009} as initialization -- we prune unused dictionary atoms on HCP dataset.

For benchmarking, we measure CPU time only, \textit{i.e.} ignore IO time as it is
very platform dependent. To limit disk access in out-of-core computation,
small memory usage is crucial for IO time.

We use \textit{scikit-learn} for computation, along with the \textit{Nilearn}
neuro-imaging library. Code for the methods and experiments is available at \href{http://github.com/arthurmensch/nilearn/tree/isbi}{\textit{http://github.com/arthurmensch/nilearn/tree/isbi}}.

\begin{figure}[t]
\vspace{-1em}
\begin{minipage}[b]{\linewidth}
  \centering
  \input{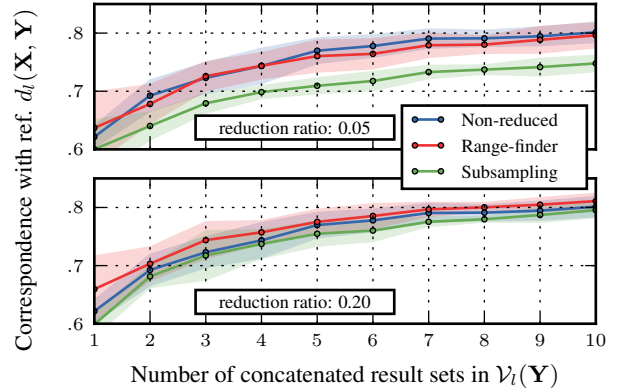}
  \vspace*{-1em}
  \caption{Result-set correspondence with non-reduced DL result-set, using different methods with different reduction ratios, increasing number of runs to show $d_l$ stabilization; variance over runs computed using 4 different result-sets $\mathcal S_l^{(0)}$; ADHD dataset.}
  \label{fig:incr_stability}
\end{minipage}
\vspace{-1.5em}
\end{figure}

\paragraph*{Indicator and reduction validity}
Fig.\;\ref{fig:incr_stability} shows $d_l$ behavior as
$l$ increases. The results demonstrate the relevance of random
range-finding as it out-performs simple subsampling. We first obtain a \textit{reference} set of maps $\mathcal V^0_p$ from non-reduced $\X$, choosing $\lambda$ to obtain little overlapping maps ($\lambda\!=\!1, 6$ for ADHD, HCP).
Secondly, we compute $d_l(\X, \mathcal S_l^0, \Y, \mathcal S_l)$ setting
$\Y = \lbrace \X, (\X_r)_{\mathrm{rf},m}, (\X_r)_{\mathrm{ss},m}
\rbrace$, for various $m \in [ n / 40, n]$. As the relationship between
$\lambda$ and a given level of sparsity
depends on $m$, we run DL on $\Y$ on a range of $\lambda$ so as to find
the value that matches best the reference run.

We observe that running DL several times does produce sets of maps that
overlap more and more, as they cover a larger part of the result-sets
$\mathcal{V}$ defined in Eq.\;\ref{eq:result_set}, and stabilizes for $l\!>\!10$. This suggest that $d_l$ does cater for randomness in DL algorithms and constitutes a good indicator for comparing two DL methods.

For $\alpha>\!.025$, and $l\!\geq\!2$,
Fig.\;\ref{fig:incr_stability} shows that compressed DL produces maps that are \textit{as comparable} with non-reduced DL maps as non-reduced DL maps obtained streaming on different orders:
\begin{equation}
d_l(\X, \mathcal S_l^0, \X, \mathcal S_l) \approx \hat d_l(\X, \mathcal S_l^0, \X_r, \mathcal S_l)
\end{equation}
Overlap between $\mathcal V_l(\X)$ and $\mathcal V_l(\X_r)$ is thus comparable to overlap between $\mathcal V_l(\X)$ and $\mathcal V_l(\X)$ for different runs from $\mathcal S_l,\, \mathcal S_l^0$.
They are therefore of the same inner quality for neuroscientists as it
not possible to tell one apart from the other.

For large compression factors -- typically with ${m\!<\!k}$, for $\alpha\!<\!.1$ on ADHD, $\alpha<\!.05$ on HCP -- range finding reduction performs significantly better than subsampling.
Both methods perform similarly for small compression factors, which shows
that subsampling already provides good \textit{large }low-rank approximation of $\X$.
Using a range-finding algorithm is therefore useful when drastically reducing data size, typically when loading very large datasets in memory.

\begin{figure}[t]
\vspace*{-1em}
\begin{minipage}[b]{\linewidth}
  \centering
  %% Creator: Matplotlib, PGF backend
%%
%% To include the figure in your LaTeX document, write
%%   \input{<filename>.pgf}
%%
%% Make sure the required packages are loaded in your preamble
%%   \usepackage{pgf}
%%
%% Figures using additional raster images can only be included by \input if
%% they are in the same directory as the main LaTeX file. For loading figures
%% from other directories you can use the `import` package
%% and then include the figures with
%%   \import{<path to file>}{<filename>.pgf}
%%
%% Matplotlib used the following preamble
%%   \usepackage[utf8x]{inputenc}\usepackage[T1]{fontenc}
%%
\begingroup%
\makeatletter%
\begin{pgfpicture}%
\pgfpathrectangle{\pgfpointorigin}{\pgfqpoint{3.149685in}{1.869000in}}%
\pgfusepath{use as bounding box, clip}%
\begin{pgfscope}%
\pgfsetbuttcap%
\pgfsetmiterjoin%
\definecolor{currentfill}{rgb}{1.000000,1.000000,1.000000}%
\pgfsetfillcolor{currentfill}%
\pgfsetlinewidth{0.000000pt}%
\definecolor{currentstroke}{rgb}{1.000000,1.000000,1.000000}%
\pgfsetstrokecolor{currentstroke}%
\pgfsetdash{}{0pt}%
\pgfpathmoveto{\pgfqpoint{0.000000in}{0.000000in}}%
\pgfpathlineto{\pgfqpoint{3.149685in}{0.000000in}}%
\pgfpathlineto{\pgfqpoint{3.149685in}{1.869000in}}%
\pgfpathlineto{\pgfqpoint{0.000000in}{1.869000in}}%
\pgfpathclose%
\pgfusepath{fill}%
\end{pgfscope}%
\begin{pgfscope}%
\pgfpathrectangle{\pgfqpoint{1.722518in}{1.040479in}}{\pgfqpoint{0.570596in}{0.480954in}} %
\pgfusepath{clip}%
\pgfpathmoveto{\pgfqpoint{1.722518in}{1.040479in}}%
\pgfpathlineto{\pgfqpoint{2.293114in}{1.040479in}}%
\pgfpathlineto{\pgfqpoint{2.293114in}{1.521434in}}%
\pgfpathlineto{\pgfqpoint{1.722518in}{1.521434in}}%
\pgfpathlineto{\pgfqpoint{1.722518in}{1.040479in}}%
\pgfusepath{clip}%
\pgftext[at=\pgfqpoint{1.720000in}{1.040000in},left,bottom]{\pgfimage[interpolate=true,width=0.580000in,height=0.490000in]{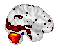}}%
\end{pgfscope}%
\begin{pgfscope}%
\pgfpathrectangle{\pgfqpoint{2.428916in}{1.002696in}}{\pgfqpoint{0.527231in}{0.556522in}} %
\pgfusepath{clip}%
\pgfpathmoveto{\pgfqpoint{2.428916in}{1.002696in}}%
\pgfpathlineto{\pgfqpoint{2.956147in}{1.002696in}}%
\pgfpathlineto{\pgfqpoint{2.956147in}{1.559217in}}%
\pgfpathlineto{\pgfqpoint{2.428916in}{1.559217in}}%
\pgfpathlineto{\pgfqpoint{2.428916in}{1.002696in}}%
\pgfusepath{clip}%
\pgftext[at=\pgfqpoint{2.430000in}{1.000000in},left,bottom]{\pgfimage[interpolate=true,width=0.540000in,height=0.570000in]{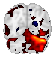}}%
\end{pgfscope}%
\begin{pgfscope}%
\pgfpathrectangle{\pgfqpoint{0.353087in}{0.317001in}}{\pgfqpoint{0.570596in}{0.480954in}} %
\pgfusepath{clip}%
\pgfpathmoveto{\pgfqpoint{0.353087in}{0.317001in}}%
\pgfpathlineto{\pgfqpoint{0.923683in}{0.317001in}}%
\pgfpathlineto{\pgfqpoint{0.923683in}{0.797955in}}%
\pgfpathlineto{\pgfqpoint{0.353087in}{0.797955in}}%
\pgfpathlineto{\pgfqpoint{0.353087in}{0.317001in}}%
\pgfusepath{clip}%
\pgftext[at=\pgfqpoint{0.350000in}{0.320000in},left,bottom]{\pgfimage[interpolate=true,width=0.580000in,height=0.490000in]{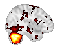}}%
\end{pgfscope}%
\begin{pgfscope}%
\pgfpathrectangle{\pgfqpoint{1.059485in}{0.279217in}}{\pgfqpoint{0.527231in}{0.556522in}} %
\pgfusepath{clip}%
\pgfpathmoveto{\pgfqpoint{1.059485in}{0.279217in}}%
\pgfpathlineto{\pgfqpoint{1.586716in}{0.279217in}}%
\pgfpathlineto{\pgfqpoint{1.586716in}{0.835739in}}%
\pgfpathlineto{\pgfqpoint{1.059485in}{0.835739in}}%
\pgfpathlineto{\pgfqpoint{1.059485in}{0.279217in}}%
\pgfusepath{clip}%
\pgftext[at=\pgfqpoint{1.060000in}{0.280000in},left,bottom]{\pgfimage[interpolate=true,width=0.540000in,height=0.570000in]{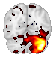}}%
\end{pgfscope}%
\begin{pgfscope}%
\pgfpathrectangle{\pgfqpoint{1.722518in}{0.317001in}}{\pgfqpoint{0.570596in}{0.480954in}} %
\pgfusepath{clip}%
\pgfpathmoveto{\pgfqpoint{1.722518in}{0.317001in}}%
\pgfpathlineto{\pgfqpoint{2.293114in}{0.317001in}}%
\pgfpathlineto{\pgfqpoint{2.293114in}{0.797955in}}%
\pgfpathlineto{\pgfqpoint{1.722518in}{0.797955in}}%
\pgfpathlineto{\pgfqpoint{1.722518in}{0.317001in}}%
\pgfusepath{clip}%
\pgftext[at=\pgfqpoint{1.720000in}{0.320000in},left,bottom]{\pgfimage[interpolate=true,width=0.580000in,height=0.490000in]{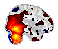}}%
\end{pgfscope}%
\begin{pgfscope}%
\pgfpathrectangle{\pgfqpoint{2.428916in}{0.279217in}}{\pgfqpoint{0.527231in}{0.556522in}} %
\pgfusepath{clip}%
\pgfpathmoveto{\pgfqpoint{2.428916in}{0.279217in}}%
\pgfpathlineto{\pgfqpoint{2.956147in}{0.279217in}}%
\pgfpathlineto{\pgfqpoint{2.956147in}{0.835739in}}%
\pgfpathlineto{\pgfqpoint{2.428916in}{0.835739in}}%
\pgfpathlineto{\pgfqpoint{2.428916in}{0.279217in}}%
\pgfusepath{clip}%
\pgftext[at=\pgfqpoint{2.430000in}{0.280000in},left,bottom]{\pgfimage[interpolate=true,width=0.540000in,height=0.570000in]{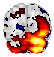}}%
\end{pgfscope}%
\begin{pgfscope}%
\pgfpathrectangle{\pgfqpoint{0.353087in}{1.040479in}}{\pgfqpoint{0.570596in}{0.480954in}} %
\pgfusepath{clip}%
\pgfpathmoveto{\pgfqpoint{0.353087in}{1.040479in}}%
\pgfpathlineto{\pgfqpoint{0.923683in}{1.040479in}}%
\pgfpathlineto{\pgfqpoint{0.923683in}{1.521434in}}%
\pgfpathlineto{\pgfqpoint{0.353087in}{1.521434in}}%
\pgfpathlineto{\pgfqpoint{0.353087in}{1.040479in}}%
\pgfusepath{clip}%
\pgftext[at=\pgfqpoint{0.350000in}{1.040000in},left,bottom]{\pgfimage[interpolate=true,width=0.580000in,height=0.490000in]{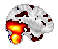}}%
\end{pgfscope}%
\begin{pgfscope}%
\pgfpathrectangle{\pgfqpoint{1.059485in}{1.002696in}}{\pgfqpoint{0.527231in}{0.556522in}} %
\pgfusepath{clip}%
\pgfpathmoveto{\pgfqpoint{1.059485in}{1.002696in}}%
\pgfpathlineto{\pgfqpoint{1.586716in}{1.002696in}}%
\pgfpathlineto{\pgfqpoint{1.586716in}{1.559217in}}%
\pgfpathlineto{\pgfqpoint{1.059485in}{1.559217in}}%
\pgfpathlineto{\pgfqpoint{1.059485in}{1.002696in}}%
\pgfusepath{clip}%
\pgftext[at=\pgfqpoint{1.060000in}{1.000000in},left,bottom]{\pgfimage[interpolate=true,width=0.540000in,height=0.570000in]{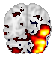}}%
\end{pgfscope}%
\begin{pgfscope}%
\pgftext[x=1.009273in,y=0.933251in,,]{\rmfamily\fontsize{9.000000}{10.800000}\selectfont Reference run}%
\end{pgfscope}%
\begin{pgfscope}%
\pgftext[x=0.245093in,y=0.904217in,left,base,rotate=90.000000]{\rmfamily\fontsize{9.000000}{10.800000}\selectfont Non-reduced \(\displaystyle \mathbf X\)}%
\end{pgfscope}%
\begin{pgfscope}%
\pgftext[x=0.638385in,y=1.656440in,,]{\rmfamily\fontsize{9.000000}{10.800000}\selectfont \(\displaystyle x = 42\)}%
\end{pgfscope}%
\begin{pgfscope}%
\pgftext[x=1.323100in,y=1.656440in,,]{\rmfamily\fontsize{9.000000}{10.800000}\selectfont \(\displaystyle z = -60\)}%
\end{pgfscope}%
\begin{pgfscope}%
\pgftext[x=2.378703in,y=0.933251in,,]{\rmfamily\fontsize{9.000000}{10.800000}\selectfont Second run}%
\end{pgfscope}%
\begin{pgfscope}%
\pgftext[x=2.007816in,y=1.656440in,,]{\rmfamily\fontsize{9.000000}{10.800000}\selectfont \(\displaystyle x = 42\)}%
\end{pgfscope}%
\begin{pgfscope}%
\pgftext[x=2.692531in,y=1.656440in,,]{\rmfamily\fontsize{9.000000}{10.800000}\selectfont \(\displaystyle z = -60\)}%
\end{pgfscope}%
\begin{pgfscope}%
\pgftext[x=1.009273in,y=0.209773in,,]{\rmfamily\fontsize{9.000000}{10.800000}\selectfont Range-finder \(\displaystyle (\mathbf X_r)_{\mathrm{rf}}\)}%
\end{pgfscope}%
\begin{pgfscope}%
\pgftext[x=0.245093in,y=0.100000in,left,base,rotate=90.000000]{\rmfamily\fontsize{9.000000}{10.800000}\selectfont Reduced \(\displaystyle \mathbf X_r\)}%
\end{pgfscope}%
\begin{pgfscope}%
\pgftext[x=2.378703in,y=0.209773in,,]{\rmfamily\fontsize{9.000000}{10.800000}\selectfont Subsampling \(\displaystyle (\mathbf X_r)_{\mathrm{ss}}\)}%
\end{pgfscope}%
\end{pgfpicture}%
\makeatother%
\endgroup%
  \vspace*{-.6em}
  \caption{Median aligned maps with various methods; HCP dataset; reduction ratio $\alpha\!=.025$.}
   \label{fig:brain}
\end{minipage}
\vspace*{-1.5em}
\end{figure}

\paragraph*{Qualitative accuracy}We validate qualitatively our results, as
this is crucial in DL decomposition: maps obtained from reduced data
should capture the same underlying neurological networks as reference
maps. In Fig.\;\ref{fig:brain}, we display matched maps when comparing
two result-sets.
For this, we find matchings between sets $(\mathcal V_l, \mathcal V_l^0)$, and
we display the maps corresponding to the median-value of this matching.
Maps are strongly alike from a neurological perspective. In particular, maps do
not differ more between our reduced dictionary learning approach and the
reference algorithm than across two runs of the reference algorithm.

\begin{figure}[t]
\vspace*{-1em}
\begin{minipage}[b]{\linewidth}
  \centering
  \input{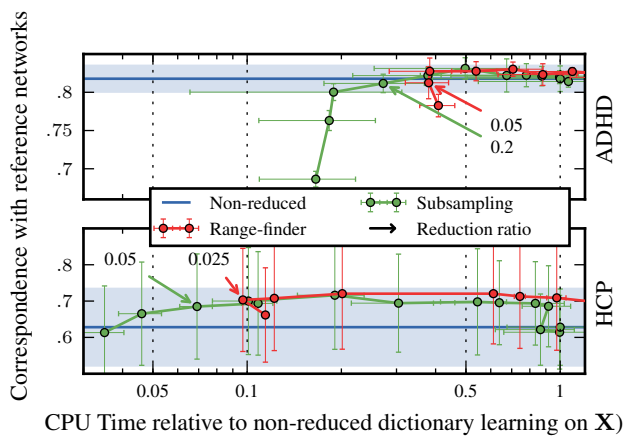}
  \vspace*{-2em}
  \caption{Time/accuracy using range-finder projectors and subsampling before DL; blue stripe recalls correspondence of results when performing different runs on non-reduced $\X$. $l\!=\!10, 3$ for ADHD, HCP. Variance over runs computed using 4 distinct subject sets $\mathcal S_l^{(0)}$.}
   \label{fig:time_v_corr}
\end{minipage}
\vspace*{-1.5em}
\end{figure}

\paragraph*{Time and accuracy tradeoff}

For efficient neuroimaging data analysis, the important quantity is the
tradeoff between quality of the results and computation time.
On Fig.\;\ref{fig:time_v_corr},
we plot $d_{l} (\X, \Y)$ -- omitting $\mathcal S_l,\mathcal S_l^0$ in notation -- against computational CPU
time, for various $\Y$. Using
range-finding algorithm and to a lesser extent time subsampling on data
before map decomposition does not significantly deteriorate results up to
large reduction factor, while allowing large gains in time \textit{and}
memory. Compression can be higher for larger datasets: we can reduce our
HCP subset up to 40 times, ADHD up to 20 times, keeping $d_l(\X, \X_r)$
within the standard deviation of $d_l(\X, \X)$.

\begin{table}[t]
\begin{center}
\begin{tabular}{|c|c|c|c|c|c|}
\hline
Data & RF $\alpha$&\multicolumn{2}{c|}{CPU Time} &
\multicolumn{2}{c|}{Corresp. $d_{l}(\X, \Y)$} \\
\hline
& & Red. & N-red. & Reduced & Non-red. \\
\hline
HCP & $.025$ & $\mathbf{849\,s}$ & $7425\,\mathrm s$ & $.703\!\pm\!.141$ & $.628\!\pm\!.105$\\
\hline
ADHD & $.05$ & $\mathbf{71\,s}$ & $186\,\mathrm s$ & $.796\!\pm\!.020$ & $.801\!\pm\!.016$\\
\hline
\end{tabular}
\caption{Time/accuracy with most interesting method for each dataset, comparing to reference DL run. RF $\alpha$: range-finder ratio}
\label{table:tradeoff}
\end{center}
\vspace*{-.8em}
\end{table}
The range-finder algorithm adds a time overhead that shift performance curve
towards higher time for large compression. However, it allows 4 times lower memory usage and thus
higher overall efficiency when considering IO. Moreover, benchmarks were
performed on a single core, while reduction can be parallelized over
subjects to reduce its overhead.

We outline best time/accuracy trade-off reduction ratios in
Fig.\;\ref{fig:time_v_corr} and Table \ref{table:tradeoff}. They depend
on chosen $k$ and on dataset, but any reasonably low reduction (with
$m\!\precsim\!k$) ratio is likely to produce good results with little
accuracy loss. Following this strategy, we set $\alpha=.025$ and
performed the entire processing of
100 subjects of the HCP dataset (384GB) on a single workstation (64GB RAM) in less than 7 hours.

%reducing HCP ten times, i.e using 40 \textit{summary} samples per subjects, allows to recover maps of similar quality as when performing DL without reduction.

%Yet, the gain in time is significant, despite range-finding overhead that shifts range-finder curve towards higher times.

\section{Acknowledgement}

The research leading to these results has received funding
from the European Union Seventh Framework Programme
(FP7/2007-2013) under grant agreement no. 604102
(Human Brain Project).

\section{Conclusion}
We introduce the use of a randomized range finding algorithm to reduce large scale datasets before performing dictionary learning and extract spatial maps. To prove efficiency of time reduction before dictionary learning, we have designed a meaningful indicator to measure result maps \textit{correspondence}, and have demonstrated that fMRI time samples have a low rank structure that allows range finding projection to be more efficient than simple subsampling.

This approach enables a 40-fold data reduction upon loading of each
subjects. It thus makes processing large datasets such as the HCP
($1.92\mathrm{TB}$) tractable on a single workstation, time-wise and
memory-wise.

\bibliographystyle{IEEEbib}
\bibliography{bibliography}

\end{document}